\documentclass[11pt]{article}

\usepackage[final]{acl}
\usepackage{times}
\usepackage{latexsym}

\usepackage[T1]{fontenc}

\usepackage[utf8]{inputenc}

\usepackage{microtype}

\usepackage{inconsolata}

\usepackage{graphicx}
\usepackage{multirow}
\usepackage{bm}
\usepackage{bbm}
\usepackage{amsmath}
\usepackage{amsthm}
\usepackage{amssymb}
\usepackage{multirow}
\usepackage{float}
\usepackage[switch]{lineno}
\usepackage{newfloat}
\usepackage{listings}
\usepackage{algorithm}
\usepackage{algorithmic}
\usepackage{xcolor}
\usepackage{booktabs}
\usepackage{subfigure}
\usepackage{dsfont}
\title{
Enhancing Reinforcement Learning for Radiology Report Generation with Evidence-aware Rewards and Self-correcting Preference Learning}

\author{
 \textbf{Qin Zhou\textsuperscript{1,2}}\thanks{These authors contributed equally.},
 \textbf{Guoyan Liang\textsuperscript{3,4}}\footnotemark[1],
 \textbf{Qianyi Yang\textsuperscript{3,4}},
 \textbf{Jingyuan Chen\textsuperscript{3,4}},
\\
 \textbf{Sai Wu\textsuperscript{3,4}}\thanks{Corresponding Authors.},
 \textbf{Chang Yao\textsuperscript{3,4}}\footnotemark[2],
 \textbf{Zhe Wang\textsuperscript{1,2}}\footnotemark[2],
\\
\\
 \textsuperscript{1}Department of Computer Science and Engineering, ECUST, China, \\
 \textsuperscript{2}Key Laboratory of Smart Manufacturing in Energy Chemical Process, Ministry of Education, P. R. China, \\
 \textsuperscript{3}Zhejiang University, Hangzhou, China, \\
 \textsuperscript{4}Hangzhou High-Tech Zone (Binjiang) Institute of Blockchain and Data Security, \\
    \small{
    guoyanl@zju.edu.cn
 }
}

\begin{document}
\maketitle
\begin{abstract}

Recent reinforcement learning (RL) approaches have advanced radiology report generation (RRG), yet two core limitations persist: (1) report-level rewards offer limited evidence-grounded guidance for clinical faithfulness; and (2) current methods lack an explicit self-improving mechanism to align with clinical preference. We introduce clinically aligned \textbf{E}vidence-aware \textbf{S}elf-\textbf{C}orrecting \textbf{R}einforcement \textbf{L}earning (ESC-RL), comprising two key components. First, a Group-wise Evidence-aware Alignment Reward (GEAR) delivers group-wise, evidence-aware feedback. GEAR reinforces consistent grounding for true positives, recovers missed findings for false negatives, and suppresses unsupported content for false positives. Second, a Self-correcting Preference Learning (SPL) strategy automatically constructs a reliable, disease-aware preference dataset from multiple noisy observations and leverages an LLM to synthesize refined reports without human supervision. ESC-RL promotes clinically faithful, disease-aligned reward and supports continual self-improvement during training. Extensive experiments on two public chest X-ray datasets demonstrate consistent gains and state-of-the-art performance.
\end{abstract}

\section{Introduction}
Radiology reports translate medical images into clinical knowledge, enabling efficient interpretation and decision-making. Yet producing high-quality reports demands careful attention to subtle visual cues and precise medical terminology, making it time-consuming and cognitively intensive. As imaging volumes surge, automated radiology report generation offers a promising path to reduce radiologists’ workload.

Current RRG approaches have achieved remarkable progress by incorporating knowledge graphs \cite{yin-etal-2025-kia}, contrastive learning \cite{li2024contrastivelearningcounterfactualexplanations}, retrieval augmentation \cite{zhou2025learnableretrievalenhancedvisualtext}, and large language models (LLMs) \cite{hou2025radarenhancingradiologyreport,zhang2025libraleveragingtemporalimages}. 
Recently, Reinforcement learning (RL) has gained traction for its strong empirical performance in gameplay, robotics, autonomous systems, and multimodal learning \cite{kaufmann2023champion,cheng2024rime}.
Inspired by these advances, RL-based RRG methods have shown promising results, largely due to carefully designed reward functions \cite{xiao2024radiologyreportgenerationmultiobjective}.
Prior works \cite{qin-song-2022-reinforced,zhou2024largemodeldrivenradiology} leverage NLG or clinical efficacy (CE) metrics to align RRG models.
However, such rewards provide limited evidence-based guidance. 
Moreover, while clinical reports require preference alignment, existing Preference-based RL (PbRL) methods \cite{xiao2024radiologyreportgenerationmultiobjective,cheng2024rime}  typically rely on report-level preference datasets and lack a self-improving mechanism to correct unreliable descriptions.  
To address these gaps, we propose an \textbf{E}vidence-aware \textbf{S}elf-\textbf{C}orrecting \textbf{R}einforcement \textbf{L}earning (ESC-RL) framework, which tackles limited evidence guidance and disease-specific self-correction in report generation.


To provide effective clinical evidence-based guidance under weak supervision, we introduce a Group-wise Evidence-aware Alignment Reward (GEAR) module.
GEAR enhances fine-grained image–report alignment via a group-wise alignment reward. It first compares disease-status vectors from ground-truth and generated reports, partitioning predictions into true positives (TPs), false negatives (FNs), and false positives (FPs). Using Disease-grounded Response Maps (DRMs), GEAR imposes group-wise evidence-aware constraints: (1) for TPs, it enforces precise grounding by maximizing IoUs between predicted and ground-truth DRMs; (2) for FNs, it aims to recover missed evidence by minimizing differences between the predicted and corresponding ground-truth DRMs; and (3) for FPs, it penalizes unsupported claims by discouraging highly activated DRMs with irrelevant regions. These designs yield disease-specific, evidence-aware RL rewards that provide clinically grounded feedback for policy optimization.


To further align generated reports with clinical preferences, we develop a Self-correcting Preference Learning (SPL) strategy.
SPL constructs a disease-specific preference dataset from multiple candidate report-level observations and uses it to train a lightweight predictor with a disease-specific description selection mechanism. Specifically, the predictor scores each disease-specific description, after which the selection mechanism identifies and filters unreliable descriptions. The retained trustworthy data then guides the re-integration of disease-specific observations to produce a more accurate, refined report.
Our contributions can be summarized as follows:
\begin{itemize}
\item We propose a novel ESC-RL framework to integrate a Group-wise Evidence-aware Alignment Reward (GEAR) and a Self-correcting Preference Learning (SPL) strategy to address limited evidence guidance and disease-specific self-correction in RL-based RRG.
\item GEAR aligns disease-specific visual–text groundings between predicted and ground-truth reports via group-wise constraints, providing disease-specific, evidence-aware feedback for policy optimization.
\item SPL identifies and selects reliable disease-level descriptions from multiple noisy generated observations, enabling accurate disease-specific description refinement.
\item Extensive experiments, including comparisons with state-of-the-art RRG methods and ablation studies, consistently demonstrate the superior performance of our approach.
\end{itemize}

\section{Related Works}
\subsection{Radiology Report Generation}
Radiology report generation (RRG) focuses on generating detailed and clinically accurate textual descriptions from medical images. 
Recent advancements in RRG have explored a variety of techniques aimed at improving the quality, relevance, and accuracy of the generated reports.
These approaches include knowledge graphs ~\cite{yin-etal-2025-kia}, contrastive learning~\cite{li2024contrastivelearningcounterfactualexplanations,Liu_2025}, retrieval-augmented methods~\cite{zhou2025learnableretrievalenhancedvisualtext}, memory alignment~\cite{chen2020generating,chen2022crossmodalmemorynetworksradiology},  reinforcement learning~\cite{qin-song-2022-reinforced}, human preference optimization~\cite{xiao2024radiologyreportgenerationmultiobjective}, and LLM-based methods~\cite{hou2025radarenhancingradiologyreport,zhang2025libraleveragingtemporalimages}.
Despite recent advances, radiology report generation (RRG) remains far from robust clinical deployment due to reliability issues, including omissions and hallucinations of critical findings, and the lack of self-improving mechanisms to align with evolving clinical preferences, keeping RRG an active research area.

\subsection{Reinforcement Learning}
Reinforcement learning (RL) is a learning paradigm for sequential decision-making, where an agent interacts with the environment and learns a policy to maximize cumulative rewards.
RL has achieved strong empirical success across diverse domains such as gaming, robotics, finance, and healthcare \cite{cheng2024rime}.
Recently, RL has been explored for report generation tasks \cite{qin-song-2022-reinforced,zhou2024largemodeldrivenradiology,xiao2024radiologyreportgenerationmultiobjective}.
For example, ~\cite{qin-song-2022-reinforced,zhou2024largemodeldrivenradiology} leverages NLG metrics such as BLEU, RadCliQ to guide cross-modal alignment between visual and textual features, using these metrics as rewards for the RL process.
~\cite{xiao2024radiologyreportgenerationmultiobjective} introduces a multi-dimensional reward framework to align the generated reports with multiple human preferences.
Despite this progress, these methods largely depend on report-level reward signals or manually designed multi-objective rewards, which provide limited evidence-level guidance and may be constrained by training-set coverage \cite{xiong2024iterativepreferencelearninghuman}.
Alternatively, some Preference-based RL (PbRL) approaches \cite{chexagent-2024,xiao2025onlineiterativeselfalignmentradiology} construct report-level preference datasets using strong foundation models and obtain preference labels via LLM-based scoring or metric-based evaluation, eliminating the requirement for manually designed reward functions.
However, these methods align only at the report level, lacking fine-grained preference alignment and an explicit self-improvement mechanism.


\begin{figure*}[t]
\centering
\includegraphics[width=0.89\textwidth]{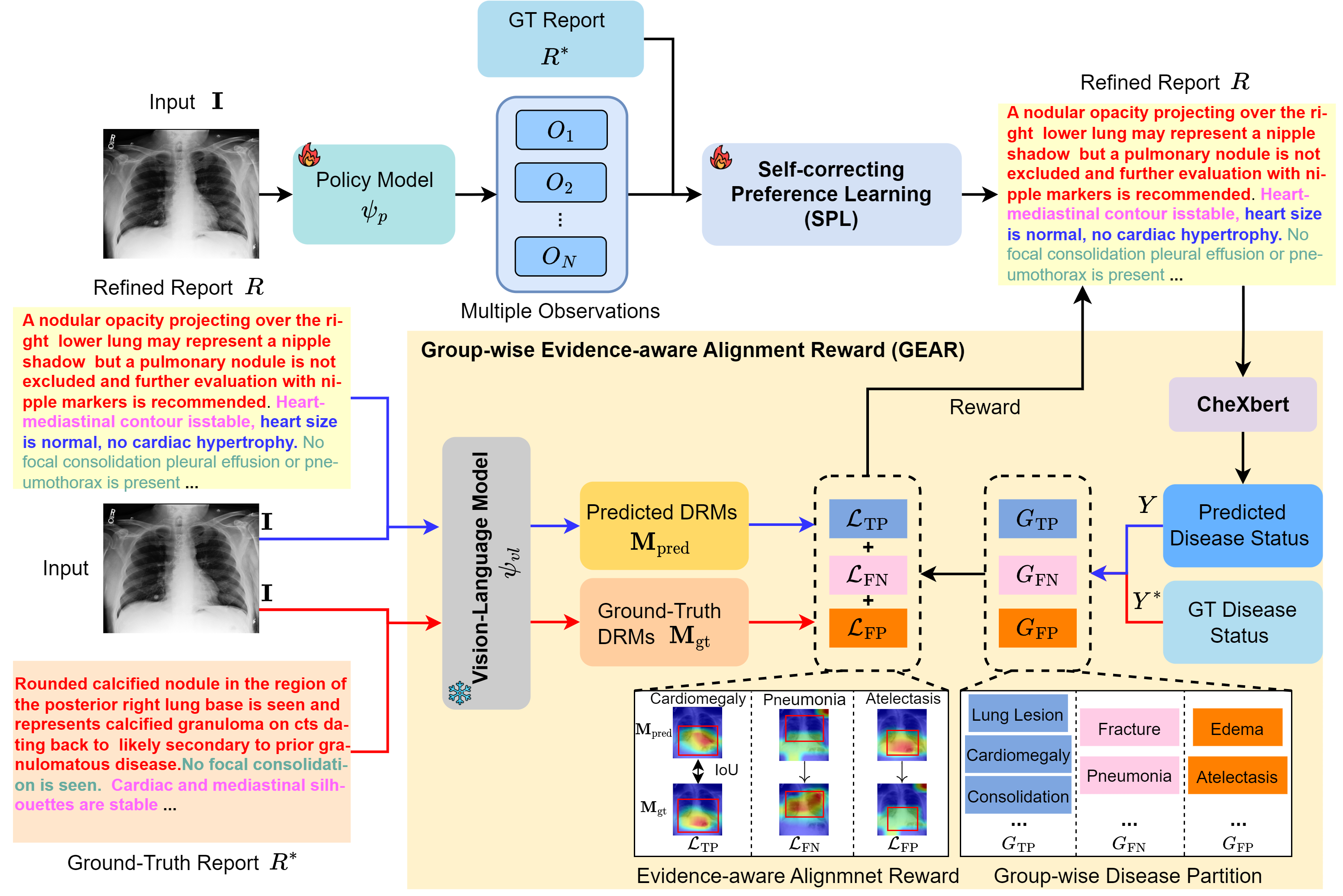}
\caption{Overview of the proposed Evidence-aware Self-Correcting Reinforcement Learning (ESC-RL) framework. }
\label{fig1:env}
\end{figure*} 

\section{Methods}

\subsection{Preliminaries}
Reinforcement learning (RL) based RRG aims to directly optimize report generation quality by treating the report generator as an agent that interacts with an environment defined by visual evidence and previously generated text. Specifically, the model parameters $\theta$ defines a policy  $\psi_{p}$ that determines the next action (i.e., the prediction of the next word).
Given a radiology image $\mathbf{I}$, the probability of generating a report $R=(y_1,\ldots,y_L)$ of length $L$ is formulated as,
\begin{equation}
  \label{eq:1}
  \psi_{p}(R|\mathbf{I}) = \prod_{l=1}^L \psi_{p}(y_{l} | y_{1}, ..., y_{l-1}, \mathbf{I}),
\end{equation}
where $\mathbb{V}$ denotes the vocabulary, $y_l \in \mathbb{V}$ is a token.

Upon generating the end-of-sequence (EOS) token, the environment returns a scalar reward $r(R)$ to evaluate the overall quality of the generated report.
RL training optimizes $\theta$ by maximizing the expected reward under the policy distribution, equivalently minimizing the negative expected reward,
\begin{equation}
  \label{eq:3}
  \mathcal{L}_{\mathrm{RL}}(\theta)=-\mathbb{E}_{R\sim \psi_{p}}[r(R)].
\end{equation}

Nevertheless, recent RL-based RRG methods \cite{qin-song-2022-reinforced,zhou2024largemodeldrivenradiology,chexagent-2024,xiao2025onlineiterativeselfalignmentradiology} still face three essential challenges: (1) the lack of fine-grained, evidence-based rewards for clinically faithful report generation; (2) the reliance on report-level preference construction for sample selection; and (3) the absence of an explicit self-correcting mechanism to support self-improvement.
The proposed ESC-RL framework is explicitly designed to overcome these limitations. Details are presented in the following.

\subsection{Framework Overview}
Figure \ref{fig1:env} illustrates the overall workflow of our ESC-RL framework. 
Given a chest X-ray image $\mathbf{I}\in\mathbb{R}^{H\times W\times 3}$, our goal is to generate a clinically accurate radiology report $R = \{y_1, y_2, ...,y_L\}$, where $(H, W)$ denotes the spatial resolution of the input image. 
Denote the ground-truth disease-status labels as $Y^*\in\{0,1,2,3\}^{K}$, where $K$ denotes the number of disease categories and $\{0,1,2,3\}$ correspond to blank, positive, negative, and uncertain status, respectively.
The image $\mathbf{I}$ is first fed into the policy model $\psi_{p}$ to sample $N$ candidate observations $\{O_n\}_{n=1}^{N}$. 
Then $\{O_n\}_{n=1}^{N}$ together with the ground-truth report $R^*$ are processed by the Self-correcting Preference Learning (SPL) module to obtain a refined report $R$.

To provide clinically-grounded evidence guidance, we incorporate a Group-wise Evidence-aware Alignment Reward (GEAR). 
GEAR compares the disease-status vectors $Y$ and $Y^*$ extracted from both the predicted and ground-truth reports using CheXbert, grouping predictions into true-positives (TPs), false-negatives (FNs), and false-positives (FPs).  Disease-grounded Response Maps (DRMs) from the predicted and ground-truth reports are then utilized to design group-wise rewards for TPs, FNs and FPs, respectively, aiming to enforce consistent DRMs for TP group, promote missed DRMs recovery for FN group, and suppress hallucinated evidence for FP group. The SPL strategy automatically scores and selects reliable  disease-specific descriptions from multiple noisy observations to generate the final refined report.


\subsection{Group-wise Evidence-aware Alignment Reward}
Fine-grained evidence is crucial for clinical diagnosis, as radiologists must localize subtle abnormalities and justify report statements with corresponding image regions.
Without explicit disease-evidence level annotations, existing RL-based approaches \cite{qin-song-2022-reinforced,zhou2024largemodeldrivenradiology} often rely on coarse report-level rewards, leading to missed findings and hallucination. 
To mitigate this issue, we propose a novel GEAR module, as shown in Figure \ref{fig1:env}.
GEAR compares Disease-grounded Response Maps (DRMs) derived from generated and ground-truth reports, and applies group-wise alignment reward to enforce region-level consistency for true-positive diseases, promote recovery for false-negative diseases, and suppress unsupported activations for false-positive diseases.

\paragraph{Group-wise Disease Partition.}
Given the refined report $R$, we use CheXbert to extract the predicted disease-status vector $Y \in \{0,1,2,3\}^K$. 
Then we compare $Y$ and the ground-truth disease-status labels $Y^*$ to form three meaningful groups: true positives (TP), false negatives (FN), and false positives (FP). It is worth noting that 
negative and uncertain status are excluded during group partition for simplicity.
Specifically, TP group $G_{\mathrm{TP}}$ contains diseases that are correctly predicted as present in both \(Y\) and \(Y^*\):
\begin{equation}
\label{eq:4}
  G_{\mathrm{TP}} = \{k, | Y^*_k =1, Y_k=1\}.
\end{equation}

FN group $G_{\mathrm{FN}}$ contains diseases which are present in $Y^*$ but missed in $Y$,
\begin{equation}
\label{eq:5}
  G_{\mathrm{FN}} = \{k, | Y^*_k =1, Y_k=0\}.
\end{equation}

FP group $G_{\mathrm{FP}}$ contains diseases that are predicted as present in \(Y\) but absent in \(Y^*\),
\begin{equation}
\label{eq:6}
  G_{\mathrm{FP}} = \{k, | Y^*_k =0, Y_k=1\}. 
\end{equation}
This decomposition allows GEAR to explicitly reinforce correctly recognized findings (TP), recover missed findings (FN) and suppress hallucinated findings (FP).

\paragraph{Disease-grounded Response Maps (DRMs) Generation.}
Given a CXR image $\mathbf{I}$, we use a frozen vision–language grounding model $\psi_{vl}$ pretrained on large-scale image–report pairs (e.g., MAVL~\cite{phan2024decomposing}) to extract disease-grounded response maps (DRMs). Specifically, we obtain predicted DRMs $\mathbf{M}^{\mathrm{pred}}$ conditioned on $\mathbf{I}$ and the generated refined report $R$, and ground-truth DRMs $\mathbf{M}^{\mathrm{gt}}$ conditioned on $\mathbf{I}$ and the ground-truth report $R^*$:
\begin{equation}
\begin{aligned}
\label{eq:7}
 &\mathbf{M}^{\mathrm{pred}} = \psi_{vl}(\mathbf{I}, R) \in \mathbb{R}^{H \times W \times K}, \\
 &\mathbf{M}^{\mathrm{gt}} = \psi_{vl}(\mathbf{I}, R^*) \in \mathbb{R}^{H \times W \times K},
\end{aligned}
\end{equation}
where $K$ is the number of diseases, and $(H, W)$ is the spatial resolution of the response map.

\paragraph{Evidence-aware Alignment Reward.}
For group $G_{\mathrm{TP}}$, we enforce spatial consistency between $\mathbf{M}^{\mathrm{pred}}$ and $\mathbf{M}^{\mathrm{gt}}$ using an IoU-based loss, which promotes consistent spatial coverage between predicted and ground-truth evidence maps,
\begin{equation}\small
  \begin{aligned}
  \label{eq:8}
      \mathcal{L}_{\mathrm{TP}} = &1 - \frac{1}{|G_{\mathrm{TP}}|}\sum _{k\in G_{\mathrm{TP}}} \\
      &\frac{2\sum\nolimits_{h,w} \mathbf{M}^{\mathrm{pred}}_{h,w,k}\mathbf{M}^{\mathrm{gt}}_{h,w,k}+\epsilon}{\sum\nolimits_{h,w}(\mathbf{M}^{\mathrm{pred}}_{h,w,k})^2 + \sum\nolimits_{h,w}(\mathbf{M}^{\mathrm{gt}}_{h,w,k})^2 +\epsilon},
  \end{aligned}
\end{equation}
where $\epsilon$ is a small constant for numerical stability.

For group $G_{\mathrm{FN}}$, we encourage the predicted DRMs to match the ground-truth DRMs via an MSE objective, thus penalizing omissions and promoting evidence recovery, 
\begin{equation}\small
  \label{eq:9}
  \mathcal{L}_{\mathrm{FN}} =  - \frac{1}{|G_{\mathrm{FN}}|}  \sum _{k\in G_{\mathrm{FN}}} \frac{1}{H W} \sum _{h,w} (\mathbf{M}^{\mathrm{pred}}_{h,w,k}-\mathbf{M}^{\mathrm{gt}}_{h,w,k})^2.
\end{equation}

For group $G_{\mathrm{FP}}$, we suppress unsupported activations by penalizing the response energy of $\mathbf{M}^{\mathrm{pred}}$, which discourages the model from producing strong visual evidence for hallucinated findings,
\begin{equation}\small
  \label{eq:10}
  \mathcal{L}_{\mathrm{FP}} = \frac{1}{|G_{\mathrm{FP}}|} \sum _{k\in G_{\mathrm{FP}}} \frac{1}{HW}  \sum _{h,w} (\mathbf{M}^{\mathrm{pred}}_{h,w,k})^2.
\end{equation}

The overall group-wise alignment reward $\mathcal{L}_{\mathrm{R}}$ combines rewards from the three groups as,
\begin{equation}
  \label{eq:11}
  \mathcal{L}_{\mathrm{R}} = \mathcal{L}_{\mathrm{TP}} + \mathcal{L}_{\mathrm{FN}} + \mathcal{L}_{\mathrm{FP}}.
\end{equation}

\begin{figure}[t]
\centering
\includegraphics[width=0.49\textwidth]{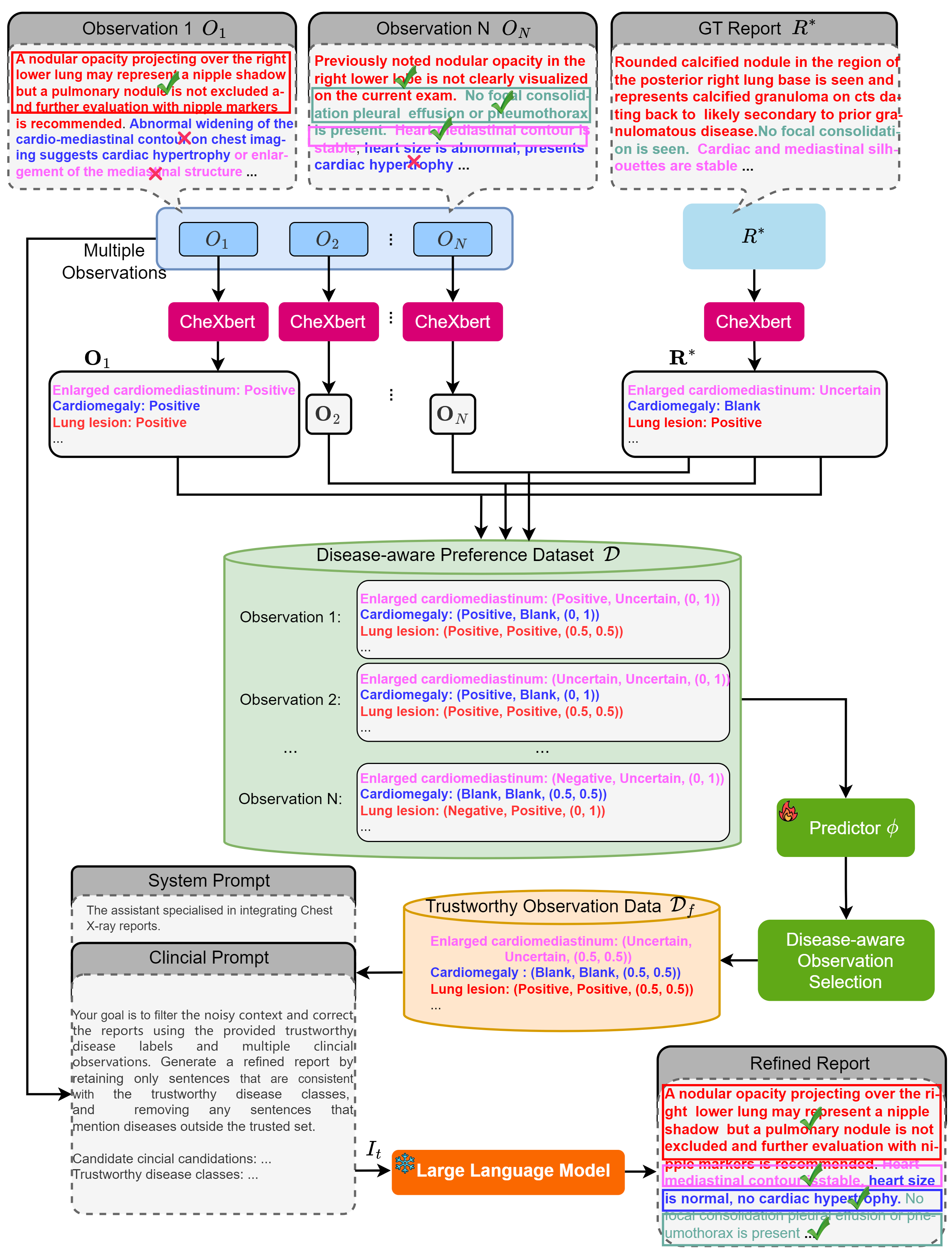}
\caption{Illustration of the Self-correcting Preference Learning (SPL) module. }
\label{fig2:env}
\end{figure} 

\subsection{Self-correcting Preference Learning}
The Self-correcting Preference Learning (SPL) strategy is designed to enable self-improvement to better align with clinical preference.
SPL constructs a disease-aware preference dataset from multiple noisy observations without human supervision, and automatically learns a predictor and selector to identify and filter unreliable disease descriptions.
The overall workflow of SPL is illustrated in Figure \ref{fig2:env}.

\paragraph{Disease-aware Preference Dataset Construction.}
Given $N$ candidate observations $\{O_n\}_{n=1}^{N}$ generated by the policy model $\psi_p$, we first apply CheXbert to extract a disease-status vector for each observation. We also extract the ground-truth disease-status vector from the target report. 
To facilitate preference learning at the disease level, we convert the disease-status vector into clinically meaningful natural-language disease descriptions.
Specifically, for the $n$-th observation, we obtain $\mathbf{O}_n = \{O_n^k\}_{k=1}^{K}$, and for the ground-truth report we obtain $\mathbf{R}^* = \{R^{*k}\}_{k=1}^{K}$, where $O_n^k$ and $R^{*k}$ denote the textual descriptions associated with the $k$-th disease.
We then construct disease-specific pairwise preference dataset in two steps:

(1) Dataset Construction. For the $k$-th disease category in observation $O_n$, we form a pair $(O_n^k,R^{*k})$, and finally result in a set of disease-wise pairs $\{(O_n^k, R^{*k})\}_{n=1,k=1}^{N,K}$;

(2) Quality Scoring. For each pair $(O_n^k, R^{*k})$, we assign a two-dimensional preference label $\tilde{o}_n^k \in\{(1,0),(0,1),(0.5,0.5)\}$ according to the relationship between $O_n^k$ and $R^{*k}$.
Specifically, $(1,0)$ indicates that $O_n^k$ is consistent with $R^{*k}$ for disease $k$, $(0,1)$ indicates that $R^{*k}$ is inconsistent with $O_n^k$ for disease $k$, and $(0.5,0.5)$ denotes an indistinguishable case where $O_n^k$ cannot be reliably judged as correct or incorrect with respect to $R^{*k}$. In our method, the preference is determined with $R^{*k}$ as the reference according to LLMs. Each disease description is then stored as a triplet $(O_n^k, R^{*k},\tilde{o}_n^k)$ in the disease-aware preference dataset $\mathcal{D}$, which is subsequently used to train the predictor model $\phi$.

\paragraph{Preference Learning.}
The continual policy updates in RL induce a non-stationary sampling distribution, leading to noisy and unstable weak preference labels.
Inspired by \cite{cheng2024rime}, we use a dual-threshold sample selector to filter and preserve reliable samples.
Concretely, for each disease-aware preference pair $(O_n^k,R^{*k})$, we use the preference predictor $\phi$ to produce a categorical distribution $\phi(O_n^k, R^{*k})$.
We measure the reliability of the pair by the divergence between the target distribution and the predicted distribution:
$D_{\mathrm{KL}}(\tilde{o}_n^k|\phi(O_n^k, R^{*k}))$.
Intuitively, pairs with large divergence are likely to be mislabeled or inconsistent and are therefore removed.
Following \cite{cheng2024rime}, we select a trustworthy subset $\mathcal{D}_f$ from the original preference dataset $\mathcal{D}$ via the lower bound $\tau_{\mathrm{lower}}$ and upper bound $\tau_{\mathrm{upper}}$ as follows,
\begin{equation}\small
 \begin{aligned}
     \mathcal{D}_{f}
  = &
  \Big\{(O^k_n,R^{*k},\tilde{o}_n^k)\ \Big|\  \\
  & \tau_{\mathrm{lower}} < D_{\mathrm{KL}}\big(\tilde{o}_n^k\ \|\ \phi(O_n^k,R^{*k})\big) < \tau_{\mathrm{upper}}\Big\},
\end{aligned}
  \label{eq:12}
\end{equation}
where $\phi(O^k_n,R^{*k})$ denote the predictor's output distribution and $\tilde{o}_n^k$ is the corresponding target distribution. $\tau_{\mathrm{upper}}$ is a time-varying threshold. 

The predictor is optimized with a standard cross-entropy objective,
\begin{equation}\small
  \mathcal{L}_{\mathrm{P}} = \mathbb{E}_{(O_n^k, R^{*k},\tilde{o}_n^k)\sim\mathcal{D}}
  \big[\mathrm{CE}\big(\phi(O_n^k,R^{*k}),\tilde{o}_n^k\big)\big].
  \label{eq:14}
\end{equation}
The filtered set $\mathcal{D}_f$ is subsequently used to guide observation selection and report re-integration. 

\paragraph{LLM-guided Self-correcting.}
Considering the excellent text generation and structured reasoning capabilities of large language models (LLMs), we employ LLM as a report re-integration component to consolidate multiple candidate observations into a refined report.
Given the candidate observations $\{O_n\}_{n=1}^N$ and the trustworthy observation data $ \mathcal{D}_f$, we construct a structured prompt $I_t$ that consists of: (i) a system prompt (ii) an instruction that enforces consistency with $\mathcal{D}_f$ and the candidate observations $\{O_n\}_{n=1}^N$. 
The prompt $I_t$ is then fed into the LLM, which removes sentences that violate $\mathcal{D}_f$, retains clinically supported descriptions, and integrates the remaining information into a coherent refined report $R$.

\subsection{Training and Inference}
The overall objective is formulated as,
\begin{equation}
    \mathcal{L} = \mathcal{L}_{\mathrm{task}} + \gamma \mathcal{L}_{\mathrm{R}} + \mathcal{L}_{\mathrm{P}},
  \label{eq:15}
\end{equation}
where $\gamma$ is a hyperparameter and set to 0.5 by default, and $\mathcal{L}_{\mathrm{task}}$ is the loss of policy model.

During inference, given an input image $\mathbf{I}$, we first sample multiple candidate reports via the policy model $\psi_p$. The SPL module then filters unreliable disease-specific predictions. Finally, using the retained trustworthy signals and candidate observations, the LLM prompted with a tailored instruction, produces the final refined report.

\section{Experiments}
In this section, we demonstrate the effectiveness of our framework through comprehensive comparisons.
Owing to space constraints, we present further visualization results and ablation experiments in the \textcolor{blue}{Appendix}.

\subsection{Datasets and Experiment Setting}
\paragraph{Datasets.}
We evaluate on two public datasets: MIMIC-CXR and IU-Xray. MIMIC-CXR \cite{johnson2019mimiccxrjpglargepubliclyavailable} contains 337,110 chest X-ray images and 227,835 corresponding reports.
We follow the standard train/val/test splits from \cite{chen2020generating,chen2022crossmodalmemorynetworksradiology}.
IU-Xray \cite{2015Preparing} contains 7,470 chest X-ray images paired with 3,955 reports, and each report corresponds to either a single frontal view or a frontal-lateral view pair.
We use the same data partition protocol as \cite{chen2020generating,chen2022crossmodalmemorynetworksradiology} for a fair comparison.
Due to the scarcity of positives for certain diseases in the IU-Xray test set, we follow \cite{jin2024promptmrg,zhou2025learnableretrievalenhancedvisualtext} to evaluate the model trained on MIMIC-CXR directly on the full IU-Xray dataset.

\paragraph{Evaluation Metrics.}
We report both lexical and radiology-specific metrics. 
For lexical evaluation, we report BLEU1, BLEU4 \cite{papineni-etal-2002-bleu}, ROUGE-L \cite{lin-2004-rouge}, and BERTScore \cite{zhang2020bertscoreevaluatingtextgeneration} to measure textual similarity and overall language quality.
For radiology-specific evaluation, we adopt RadCliQ \cite{Yu2022.08.30.22279318}, RadGraphF1 \cite{jain2021radgraphextractingclinicalentities}, CheXbertF1 \cite{smit2020chexbertcombiningautomaticlabelers}, and GREEN \cite{Ostmeier_2024}. For all metrics except RadCliQ, higher scores indicate better performance.
We also assess disease-level clinical efficacy (CE) via precision, recall, and F1, using CheXbert to map reports to 14 disease labels.

\paragraph{Implementation Details.}
We employ pre-trained REVTAF \cite{zhou2025learnableretrievalenhancedvisualtext} as the policy model and frozen MAVL \cite{phan2024decomposing} for extracting Disease-grounded Response Maps from 224-resized images. 
The lightweight predictor consists of a Bert-base encoder\cite{devlin-etal-2019-bert} that encodes the concatenated text, followed by a fully connected classification head that maps the [CLS] representation to two logits and outputs a 2-way preference distribution.
We use GPT-5 as the integration LLM and preference determination.
The number of candidate observations $N$ is set to 4.
Filtering thresholds are set to $\tau_{\mathrm{lower}} = 3\ln{(10)}$ with a decay rate $\frac{1}{30}$, and a time-varying adaptive $\tau_{\mathrm{upper}}$ following \cite{cheng2024rime}.
We optimize with AdamW (weight decay 0.05), an initial learning rate of $5e - 5$, and a cosine learning rate schedule. We train for 6 epochs with a batch size of 18. All experiments are conducted on an NVIDIA A800 GPU (80GB) for about 20 hours using Python 3.10, PyTorch 2.4.0, and Ubuntu 22.04.

\begin{table*}
\vspace{-0.5em}
  \centering
  \scalebox{0.69}{
  \begin{tabular}{l|l|cccc|cccc}
    \hline
    \multirow{2}{*}{\textbf{Model}} & \multirow{2}{*}{\textbf{Year}}&\multicolumn{4}{c|}{\textbf{Lexical Metrics}}  & \multicolumn{4}{c}{\textbf{Radiology Metrics}}  \\
    \cline{ 3-10 }
                & & BLEU-1 $\uparrow$ &BLEU-4 $\uparrow$ & ROUGE $\uparrow$ & BERTScore $\uparrow$ & RadCliQ $\downarrow$ & RadGraphF1 $\uparrow$ & ChexbertF1 $\uparrow$ &  GREEN $\uparrow$  \\
    \hline
     R2Gen   & ACL 2020 &0.353 &0.103 &0.277 &0.886 & 2.89 & 0.195 &0.276 & 0.306   \\ 
     R2GenCMN & ACL 2021 & 0.353 & 0.106 & 0.278 & 0.867 & 2.87 & 0.199 & 0.278 & 0.308 \\
     RGRG & CVPR 2023 &  0.373 & 0.126 & 0.264 & 0.873 & 2.85 & 0.221 &0.447 & 0.313 \\
     MiniGPT-Med&  - & 0.191 & 0.012 & -& 0.636 & 2.95 & 0.164 & 0.172 & 0.211 \\
     PromptMRG & AAAI 2024 & 0.398 & 0.112 & 0.268 & 0.857& 2.77 & 0.227 & 0.476 & 0.289 \\
     MedVersa & -     & 0.280 & 0.090& - & 0.711 & \textcolor{blue}{2.45} & \textcolor{blue}{0.289} & 0.471 & \textcolor{blue}{0.381}  \\
     REVTAF & ICCV 2025 & \textcolor{blue}{0.465} & \textcolor{blue}{0.182} & \textcolor{blue}{0.336} & \textcolor{blue}{0.887} & 2.48 & 0.279 & \textcolor{blue}{0.592} & 0.344             \\
     \hline
     R2GenRL & ACL 2022 &0.381 & 0.109 & 0.287 & 0.871 & 2.83 & 0.214 & 0.278 & 0.315  \\
     CheXagent & AAAI 2024 &0.172 & 0.021 &- & 0.669 & 2.88 & 0.19 & 0.265 & 0.268 \\
     MPO & AAAI 2025 & 0.416 & 0.139 & 0.309 & 0.878 & 2.63 & 0.257 & 0.353 & 0.324 \\
     OISA & ACL 2025 & 0.428 & 0.129 & - &0.885 &2.54 & 0.273 & 0.516 & 0.341   \\
    \hline
    ESC-RL (Ours)     &- & \textbf{0.487} & \textbf{0.199} & \textbf{0.352} & \textbf{0.898} & \textbf{2.39} & \textbf{0.304} & \textbf{0.608} & \textbf{0.394}    \\
    \hline
  \end{tabular}}
  \caption{\label{tab1}
    Comparison with existing RRG methods on the MIMIC-CXR dataset. The best and second-best results are highlighted in \textbf{bold} and \textcolor{blue}{blue}, respectively. ‘-’ indicates not reported.
  }
\end{table*}

\begin{table*}
  \centering
  \scalebox{0.69}{
  \begin{tabular}{l|l|cccc|cccc}
    \hline
    \multirow{2}{*}{\textbf{Model}} & \multirow{2}{*}{\textbf{Year}}&\multicolumn{4}{c|}{\textbf{Lexical Metrics}}  & \multicolumn{4}{c}{\textbf{Radiology Metrics}}  \\
    \cline{ 3-10 }
                & & BLEU-1 $\uparrow$ &BLEU-4 $\uparrow$ & ROUGE $\uparrow$ & BERTScore $\uparrow$ &RadCliQ $\downarrow$ & RadGraphF1 $\uparrow$ & CheXbertF1 $\uparrow$ &  GREEN $\uparrow$  \\
    \hline
     R2Gen   &ACL 2020 &0.289 &0.052 &0.243 & 0.861 & 2.79 & 0.187 & 0.145 & 0.482   \\ 
     R2GenCMN & ACL 2021 & 0.289 & 0.055 & 0.246 & 0.864 & 2.78 & 0.190 & 0.147 &0.483 \\
     RGRG & CVPR 2023 & 0.266 & 0.063 &0.180 & 0.867 & 2.71 & 0.189 & 0.180 & 0.481  \\
    PromptMRG & AAAI 2024 & 0.401 &0.098& 0.281& 0.871 & 2.60 & 0.274 & 0.211 & 0.457 \\
    MedVersa & - & 0.247 & 0.047 & - & 0.884 & 2.71 & 0.209 & 0.217 & 0.516  \\
    REVTAF&ICCV2025& \textcolor{blue}{0.420} & \textcolor{blue}{0.107} & \textcolor{blue}{0.309} &\textcolor{blue}{0.886} & \textcolor{blue}{2.54} &\textcolor{blue}{0.287} & \textcolor{blue}{0.273} &\textcolor{blue}{0.522}\\
    \hline
    R2GenRL & ACL 2021 & 0.290 & 0.054 & 0.248 & 0.865 & 2.78 & 0.192 & 0.151 & 0.487  \\
    CheXagent & AAAI 2024 & 0.191 & 0.036 & -& 0.876 & 2.81 & 0.184 & 0.097 & 0.407 \\
    \hline
    ESC-RL (Ours)     &   - & \textbf{0.439} & \textbf{0.118}&\textbf{0.323} & \textbf{0.890} & \textbf{2.48} & \textbf{0.307} &\textbf{0.295} & \textbf{0.537}                     \\
    \hline
  \end{tabular}}
  \caption{\label{tab2}
            Comparison with existing RRG methods on the IU-Xray dataset. The best and second-best results are highlighted in \textbf{bold} and \textcolor{blue}{blue}, respectively. ‘-’ indicates not reported.
  }
  \vspace{-1em}
\end{table*}

\subsection{Comparison with State-of-the-Arts}

\paragraph{Quantitative Results.} We compare ESC-RL with representative RRG methods, including the \textbf{traditional} methods R2Gen \cite{chen2020generating}, R2GenCMN \cite{chen2022crossmodalmemorynetworksradiology}, RGRG \cite{Tanida_2023}, MiniGPT-Med \cite{alkhaldi2024minigptmedlargelanguagemodel}, PromptMRG \cite{jin2024promptmrg}, MedVersa \cite{zhou2025medversageneralistfoundationmodel}, REVTAF \cite{zhou2025learnableretrievalenhancedvisualtext}, as well as \textbf{RL-based }approaches R2GenRL \cite{qin-song-2022-reinforced}, CheXagent \cite{chexagent-2024}, MPO \cite{xiao2024radiologyreportgenerationmultiobjective}, OISA \cite{xiao2025onlineiterativeselfalignmentradiology}.
Detailed results on MIMIC-CXR and IU-Xray are reported in Tables \ref{tab1},\ref{tab2}, and \ref{tab3}.
On MIMIC-CXR dataset, as shown in Table \ref{tab1}, our method achieves SOTA performance on both lexical and radiology-specific metrics, consistently exceeding the second-best method. Concretely, our method obtains the absolute gains of 2.2\%, 1.7\%, 1.6\%, 1.1\%, 6\%, 1.5\%, 1.6\%, and 1.3\% over runner-up across the evaluated metrics. 
On the IU-Xray dataset, we follow \cite{jin2024promptmrg,zhou2025learnableretrievalenhancedvisualtext} to directly evaluate the full dataset using the model pretrained on MIMIC-CXR dataset. As illustrated in Table \ref{tab2}, our method delivers the best overall performance across all lexical and radiology-specific metrics, outperforming the second-best method REVTAF by 1.9\%, 1.1\%, 1.4\%, 0.4\%, 6\%, 2.0\%, 2.2\%, and 1.5\%, respectively.
For CE metrics, as shown in Table \ref{tab3}, our ESC-RL consistently surpasses the second-best method REVTAF. Specifically, on the MIMIC-CXR dataset, it achieves gains of by 0.4\%, 1.2\%, and 1.6\% in Precision, Recall, and F1, respectively. On the IU-Xray dataset, the corresponding improvements are 1.3\%, 2.9\%, and 2.2\%. Overall, ESC-RL consistently outperforms the second-best method across all reported metrics on both datasets.


\begin{figure*}[t]
\centering
\includegraphics[width=0.97\textwidth]{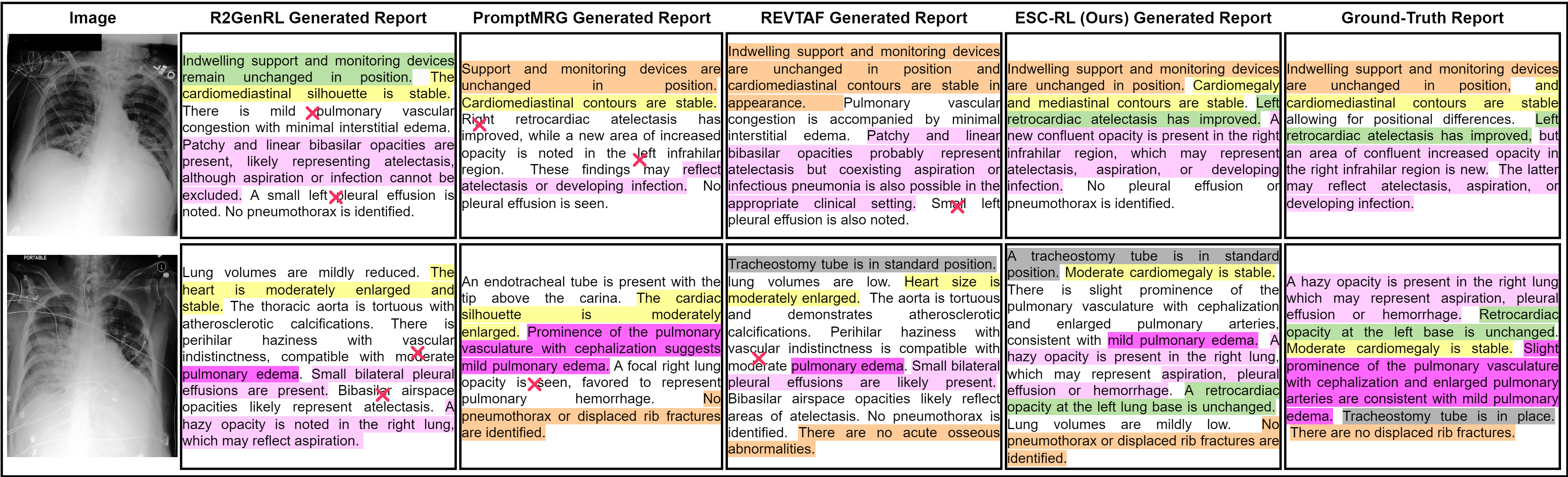}
\caption{Qualitative comparison of reports generated by R2GenRL, PromptMRG, REVTAF, and our method. Sentences are color-coded to indicate their corresponding descriptions. Incorrect statements are marked with $\times$, and unmarked sentences denote descriptions not mentioned in the ground-truth report.}
\label{fig3:env}
\vspace{-0.5em}
\end{figure*} 

\begin{table}[htb]
  \centering
  \scalebox{0.65}{
  \begin{tabular}{l|ccc|ccc}
    \hline
    \multirow{2}{*}{\textbf{Model}} &
    \multicolumn{3}{c|}{\textbf{MIMIC-CXR}}&\multicolumn{3}{c}{\textbf{IU-Xray}}  \\
    \cline{ 2-7 }
                &Precision &Recall &F1&Precision &Recall &F1   \\
    \hline
     R2Gen          &0.333 &0.273 &0.276 &0.151 & 0.145 & 0.145  \\ 
     R2GenCMN & 0.334 & 0.275 & 0.278 & 0.154 & 0.147 & 0.147  \\
     RGRG &0.461 & 0.475 & 0.447 & 0.183 & 0.187 & 0.180 \\
    PromptMRG & 0.501& 0.509& 0.476 & 0.213 & 0.229 & 0.211 \\ 
    REVTAF  &\textcolor{blue}{0.628} & \textcolor{blue}{0.613} & \textcolor{blue}{0.592} & \textcolor{blue}{0.286} & \textcolor{blue}{0.282} & \textcolor{blue}{0.273}\\
    \hline
    R2GenRL & 0.334 & 0.275 & 0.278 & 0.153 & 0.150 & 0.151 \\
    MPO & 0.436 & 0.376 & 0.353 & - & - & - \\
    \hline
    ESC-RL (Ours)    &  \textbf{0.632} & \textbf{0.625} & \textbf{0.608} &  \textbf{0.299} & \textbf{0.311} & \textbf{0.295}                     \\
    \hline
  \end{tabular}}
  \caption{\label{tab3}
    Clinical Efficacy (CE) comparison of 14 diseases on the MIMIC-CXR and IU-Xray datasets. Best values are highlighted in bold and second-best in \textcolor{blue}{blue}.
  }
\end{table}

\begin{table*}
  \centering
  \scalebox{0.67}{
  \begin{tabular}{c|lll|cccc|cccc}
    \hline
    \multirow{2}{*}{\textbf{Setting}} & \multirow{2}{*}{\textbf{RL}} & \multirow{2}{*}{\textbf{GEAR}}&\multirow{2}{*}{\textbf{SPL}}&\multicolumn{4}{c|}{\textbf{Lexical Metrics}}  & \multicolumn{4}{c}{\textbf{Radiology Metrics}}  \\
    \cline{ 5-12 }
     & &  & & BLEU-1 $\uparrow$ &BLEU-4 $\uparrow$ & ROUGE $\uparrow$ & BERTScore $\uparrow$ & RadCliQ $\downarrow$ & RadGraphF1  $\uparrow$ & CheXbertF1 $\uparrow$ & GREEN $\uparrow$   \\
    \hline
     Baseline   &  & & &  0.465 & 0.182 & 0.336 & 0.887 & 2.48 & 0.279 & 0.514 & 0.344           \\  
     \hline
     
    (a)  &  $\checkmark$ & & & 0.472 & 0.187 & 0.335 & 0.889 & 2.45 & 0.282 & 0.517 & 0.349 \\
    (b)  & & $\checkmark$ & & 0.471 & 0.191 & 0.334  & 0.891 & 2.42 & 0.285 & 0.518 & 0.358 \\
    (c)  & & & $\checkmark$ & 0.475 & 0.190 & 0.339  & 0.892 & 2.40 & 0.288 & 0.517 & 0.356\\ 
    (d)  & $\checkmark$ & $\checkmark$ & & 0.481 & 0.194 & 0.345 & 0.890 & 2.40 & 0.289 & 0.517 & 0.359 \\
    (e)  & $\checkmark$ & & $\checkmark$ & 0.483 & 0.195 & 0.344 & 0.892 & 2.40 & 0.297 & 0.519 & 0.364 \\
    (f)  & &$\checkmark$ & $\checkmark$ & 0.485 & 0.197 & 0.347 & 0.895 & 2.39 & 0.299 & 0.520 & 0.387\\
    \hline
    (g)  &$\checkmark$ & $\checkmark$&  $\checkmark$& \textbf{0.487} & \textbf{0.199} & \textbf{0.352} & \textbf{0.898} & \textbf{2.39} & \textbf{0.304} & \textbf{0.608} & \textbf{0.394} \\
    \hline
  \end{tabular}}
  \caption{\label{tab4}
            Effectiveness analysis of each component on MIMIC-CXR test set.
  }
\end{table*}
\paragraph{Qualitative Results.}
Figure \ref{fig3:env} presents two qualitative examples highlighting the superiority of ESC-RL over SOTA methods. 
As observed, ESC-RL recovers key report content, accurately identifying support devices and major findings such as cardiomediastinal contours, cardiomegaly, and atelectasis.
Moreover, ESC-RL captures fine-grained, location-specific abnormalities, e.g., left retrocardiac atelectasis and confluent opacity in the right infrahilar region. 
For ambiguous signs such as a ‘hazy right lung opacity’, ESC-RL offers clinically plausible differentials:   ‘aspiration, pleural effusion, or hemorrhage’, consistent with the ground-truth.
In contrast, R2GenRL, PromptMRG, and REVTAF often yield factual errors, incomplete descriptions, and insufficient interpretations. Overall, ESC-RL better suppresses hallucinated content and generates more accurate clinical findings, further validating the effectiveness of the proposed framework.

\begin{figure}[t]
\centering
\includegraphics[width=0.47\textwidth]{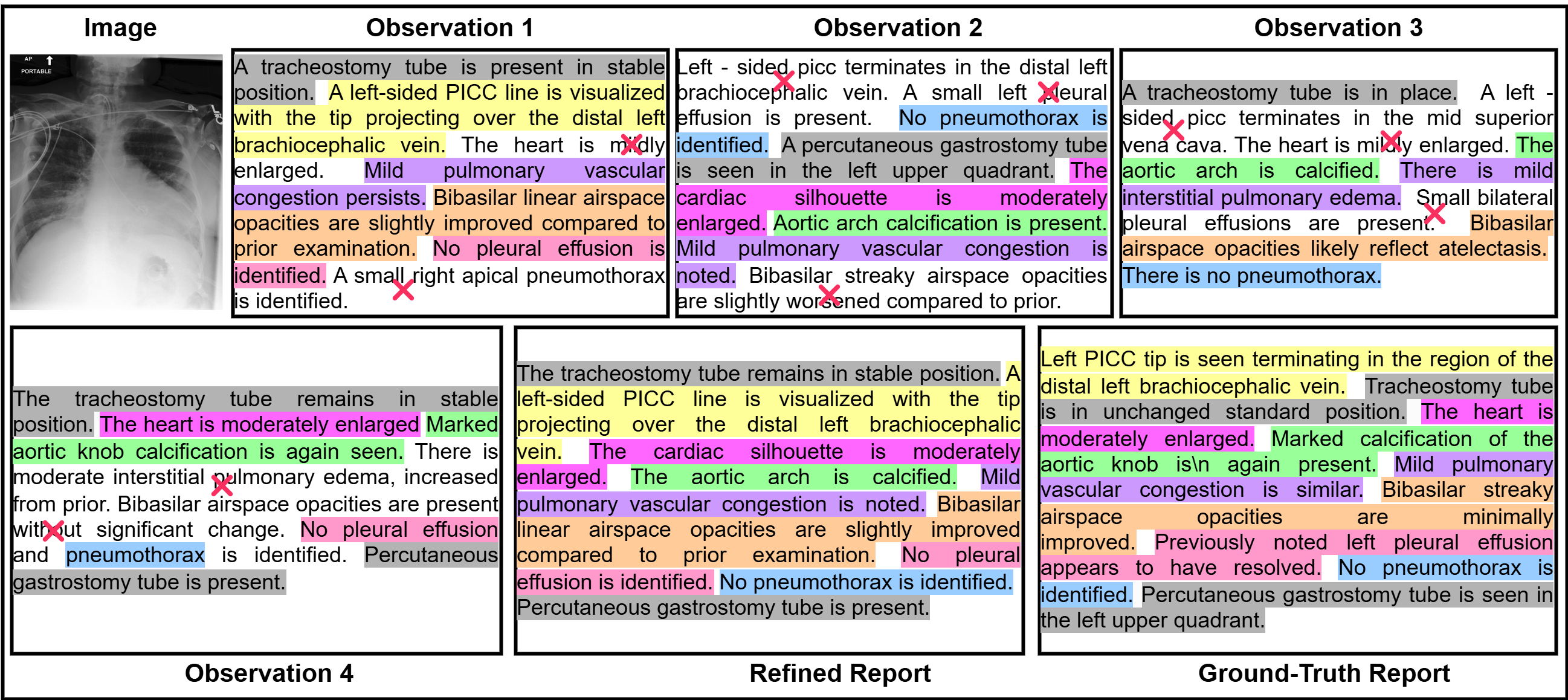}
\caption{An example of re-integrating multiple observations into a refined report. Sentences are color-coded to indicate their corresponding descriptions. Incorrect statements are highlighted with $\times$, and unmarked sentences denote content that is not mentioned in the ground-truth report.}
\label{fig4:env}
\vspace{-1em}
\end{figure} 

\subsection{Ablation Study}
We validate the effectiveness of each component in our model on the MIMIC-CXR test set, as shown in Table \ref{tab4}. As observed, 
introducing the RL framework, GEAR module, or SPL consistently improves performance over the baseline. RL alone improves GREEN by 0.5\%, indicating that report-level rewards effectively enhance cross-modal alignment. GEAR alone boosts GREEN by 1.4\%, supporting the benefit of evidence-aware reward shaping for clinically aligned grounding.
SPL alone yields a 1.2\% gain in GREEN, demonstrating the effectiveness of self-correcting preference learning.
Moreover, based on RL, introducing either GEAR or SPL yields complementary gains beyond report-level rewards.
Ultimately, integrating all proposed components, our model achieves consistent gains across all metrics, improving BLEU-1, RadGraphF1, and GREEN by 2.2\%, 2.5\%, and 5.0\%, respectively. These results highlight indispensable role of these modules in achieving stronger evidence-based clinical alignment and accurate report generation.

Figure \ref{fig4:env} illustrates a representative case where SPL re-integrates multiple intermediate observations into a refined report. As shown, observations 1–4 still contain several factual inaccuracies. After applying SPL, these errors are largely corrected, producing a refined report that better matches the ground-truth in both factual consistency and content coverage. This further validates the effectiveness of the proposed SPL strategy.

\section{Conclusion}
We propose a novel clinically aligned \textbf{E}vidence-aware \textbf{S}elf-\textbf{C}orrecting \textbf{R}einforcement \textbf{L}earning (ESC-RL) framework for RRG. ESC-RL introduces a Group-wise Evidence-aware Alignment Reward (GEAR) module that compares disease-status vectors from ground-truth and generated reports, and groups findings into TPs, FNs, and FPs. Using DRMs, GAER enforces consistent grounding for TPs, encourages recovery of missed evidence for FNs, and suppresses hallucinated evidence for FPs.  
Additionally, 
ESC-RL incorporates a Self-correcting Preference Learning (SPL) strategy that constructs a disease-aware preference dataset and uses it to train a lightweight predictor with disease-specific description selection mechanism.
SPL filters unreliable disease descriptions and leverages the retained trustworthy data to guide the observations re-integration.
Extensive experiments on two public chest X-ray datasets demonstrate consistent gains and state-of-the-art performance.

\section{Limitations}
Our experiment mainly focus on chest X-ray datasets, since they provide large-scale paired images and high-quality reports. Therefore, the generalization of ESC-RL to other modalities (e.g., CT/MRI) or anatomical regions remains to be validated. In addition, our framework relies on pretrained model to extract disease label and response maps which may introduces additional computational overhead.
Future work will extend ESC-RL toward a more modality-agnostic and unified RL framework across diverse imaging settings.

\section{Ethical Considerations}
Our study uses real-world patient data from the MIMIC-CXR and IU-Xray datasets. These datasets are de-identified and released under controlled access for research purposes. Therefore, the risk of privacy leakage is minimal. We follow the corresponding data use agreements and use the data solely for developing and evaluating automated RRG models.

\section{Acknowledgment}
This study was supported under the Key Research and Development Project of Yunnan Province (Grant No.202402AD080006) and Ningbo "Yongjiang Talent Program" Youth Innovation Project (2024A-156-G). It was also funded by the National Science Foundation of China (Grant No.62201341)

\bibliography{main}

\appendix
\section{Cost/Latency Analysis}
On our setup, the inference takes about 2.5 s/sample, and training takes about 20 h per run on 2 GPUs.
CheXbert is used only during training and is not invoked at test-time, so it does not affect deployment latency. At inference, only adding MAVL increases parameters by about 0.58M and adds about 0.3 s/sample, while alone adding GPT-5 adds about 0.7 s/sample without increasing parameters. Using both adds 0.58M and about 1.1 s/sample. These external models (MAVL, GPT-5) are fixed backbones (grounding, refinement) in the radiology report generation pipeline, which helps to moderately enhance the baseline performance, whereas ESC-RL adds minimal inference overhead while providing the main performance gains.

\section{Supplementary Ablation Studies}
\label{sec:appendix}

\begin{table*}
  \centering
  \scalebox{0.78}{
  \begin{tabular}{l|cccc|cccc}
    \hline
    \multirow{2}{*}{\textbf{Model}} &\multicolumn{4}{c|}{\textbf{Lexical Metrics}}  & \multicolumn{4}{c}{\textbf{Radiology Metrics}}  \\
    \cline{ 2-9 }
                & BLEU-1 $\uparrow$ &BLEU-4 $\uparrow$ & ROUGE $\uparrow$ & BERTScore $\uparrow$ &RadCliQ $\downarrow$ & RadGraphF1 $\uparrow$ & CheXbertF1 $\uparrow$ &  GREEN $\uparrow$  \\
    \hline
    Baseline & 0.465 & 0.182 & 0.336 & 0.887 & 2.48 & 0.279 & 0.514 & 0.344             \\
    \hline
    MedKLIP & 0.485 & \textbf{0.199} & 0.349 & 0.897 & \textbf{2.39} & 0.301 & 0.521 & 0.392\\
    \hline
    MAVL  & \textbf{0.487} & \textbf{0.199} & \textbf{0.352} & \textbf{0.898} & \textbf{2.39} & \textbf{0.304} & \textbf{0.608} & \textbf{0.394}                     \\
    \hline
  \end{tabular}}
  \caption{\label{tab5}
            Influence of different vision–language grounding models for ESC-RL on the MIMIC-CXR test set. The best results are highlighted in bold.
  }
\end{table*}
\begin{table*}
  \centering
  \scalebox{0.65}{
  \begin{tabular}{l|cccc|cccc}
    \hline
    \multirow{2}{*}{\textbf{Model}} &\multicolumn{4}{c|}{\textbf{Lexical Metrics}}  & \multicolumn{4}{c}{\textbf{Radiology Metrics}}  \\
    \cline{ 2-9 }
                & BLEU-1 $\uparrow$ &BLEU-4 $\uparrow$ & ROUGE $\uparrow$ & BERTScore $\uparrow$ &RadCliQ $\downarrow$ & RadGraphF1 $\uparrow$ & CheXbertF1 $\uparrow$ &  GREEN $\uparrow$  \\
    \hline
    TP-MSE \& FN-MSE & 0.456 & 0.177 & 0.346 & 0.859 & 2.53 & 0.264 & 0.591 & 0.335               \\
    \hline
    TP-IoU-based \& FN-IoU-based & 0.451 & 0.171 & 0.342 & 0.865 & 2.59 & 0.254 & 0.590& 0.327\\
    \hline
    TP-IoU-based \&FN-MSE (Ours)	  & \textbf{0.487} & \textbf{0.199} & \textbf{0.352} & \textbf{0.898} & \textbf{2.39} & \textbf{0.304} & \textbf{0.608} & \textbf{0.394}                     \\
    \hline
  \end{tabular}}
  \caption{\label{tab6}
            Influence of different TP/FN/FP loss for ESC-RL on the MIMIC-CXR test set. The best results are highlighted in bold.
  }
\end{table*}

\subsection{Influence of the Different Reward Weight $\gamma$}
\begin{figure}
\centering
\includegraphics[width=0.47\textwidth]{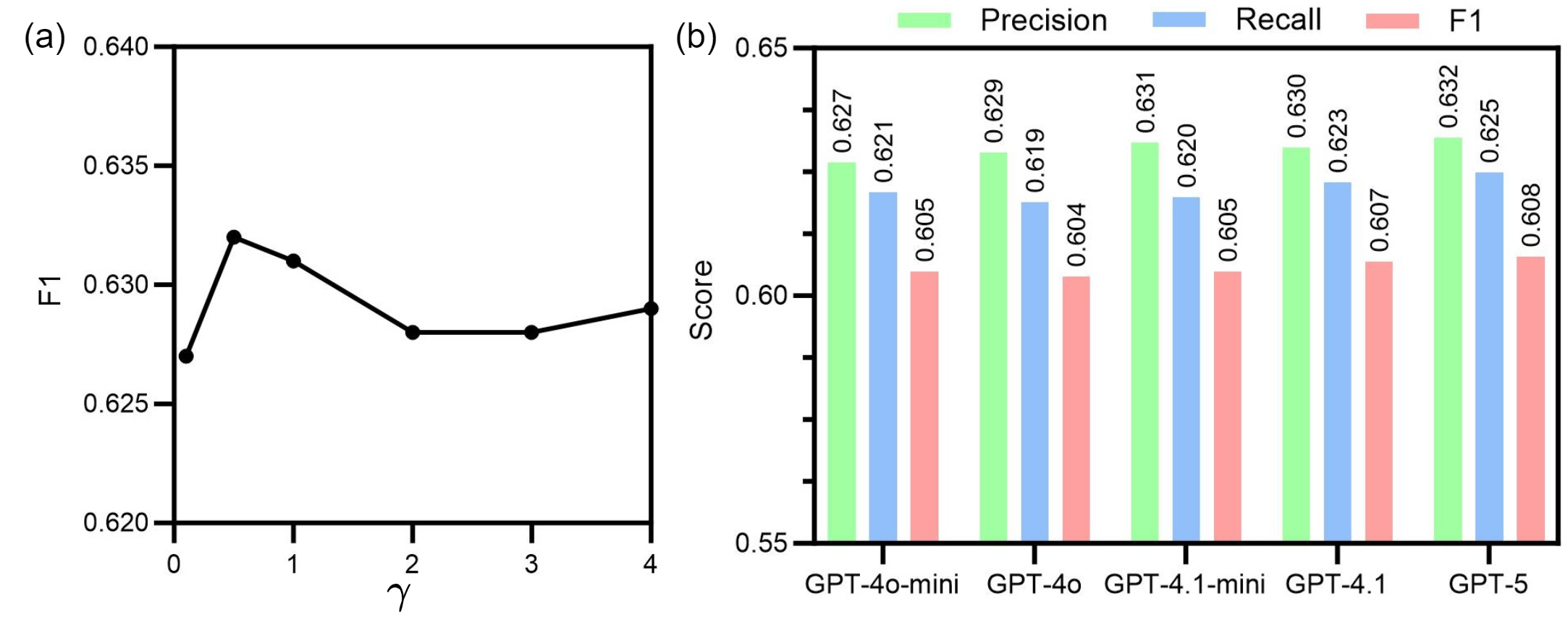}
\caption{Ablation study. (a) The influence of the different reward weight $\gamma$. (b) The influence of different LLMs for report re-integration.}
\label{fig5:env}
\end{figure}

We analyze the effect of the reward weight $\gamma$ in Figure \ref{fig5:env} (a). The results demonstrate that our model achieves the best performance when $\gamma =0.5$. In contrast, both overly large and overly small values lead to degraded outcomes. 
Intuitively, a large $\gamma$ overemphasizes the evidence-aware alignment reward, which can dominate the optimization signal and destabilize training. Conversely, a small $\gamma$ weakens the contribution of evidence-based alignment, providing insufficient guidance for clinically faithful grounding.
Overall, a moderate $\gamma$ achieves the most favorable performance, effectively balancing alignment supervision and report generation accuracy.

\subsection{Evaluation of Different LLMs for Report Re-integration}
To assess the impact of different GPT-series LLMs on report re-integration, we conduct an in-depth analysis in Figure \ref{fig5:env} (b). We evaluate several mainstream models, including GPT-4o-mini, GPT-4o, GPT-4.1-mini, GPT-4.1, and GPT-5. The results indicate that our framework consistently yields stable performance improvements across all LLMs,demonstrating the robustness of the proposed re-integration strategy.
Among the evaluated models, GPT-5 achieves the best overall performance, slightly surpassing the second-best GPT-4.1 by absolute gains of 0.2\%, 0.2\%, and 0.1\% in Precision, Recall, and F1, respectively.
This advantage is consistent with the expectation that stronger LLMs exhibit better reasoning and information aggregation capabilities, leading to more accurate refinement. 
Therefore, we adopt GPT-5 as the default re-integration model of disease-specific observations in our framework.

\subsection{Influence of Different Vision-Language Grounding Models}
To investigate the influence of the vision–language grounding model on our framework, we conduct an ablation study by replacing MAVL \cite{phan2024decomposing} with MedKLIP \cite{wu2023medklip}, as reported in Table \ref{tab5}. 
The results show that ESC-RL consistently improves over the baseline under both grounding models, demonstrating its ability to effectively exploit grounding signals for performance gains. Moreover, the stronger grounding model (MAVL) achieves slightly better results than MedKLIP, suggesting that more accurate vision-language alignment provides higher-quality supervision for evidence-aware optimization. Importantly, ESC-RL remains superior to SOTA methods regardless of the grounding model employed, indicating that our framework is not tied to a specific grounding backbone and generalizes well across different vision–language grounding models.

\subsection{TP/FN/FP Loss Selection}
Due to different supervision objectives for TP/FN/FP, we do not use a single MSE loss for Eq.\ref{eq:8}, \ref{eq:9}, and \ref{eq:10}. FP (Eq.\ref{eq:10}) is a suppression case (disease absent in $Y^*$), so there is no positive DRM to match. Overlap losses with an empty GT mask can degenerate and drive trivial collapse, hence we penalize DRM area to suppress spurious evidence. For TP, FN (Eqs.\ref{eq:8} and \ref{eq:9}), Table \ref{tab6} shows that forcing MSE loss for both or IoU-based loss for both degrades performance. MSE loss is background-dominated and blurs TP boundaries, while IoU-based loss can be unstable when FN masks are weak. Overall, the best configuration is IoU-based loss for TP, MSE loss for FN, and suppression for FP, aligned with the correct supervision semantics rather than heuristic choice.

\begin{table*}
  \centering
  \scalebox{0.75}{
  \begin{tabular}{l|cccc|cccc}
    \hline
    \multirow{2}{*}{\textbf{Model}} &\multicolumn{4}{c|}{\textbf{Lexical Metrics}}  & \multicolumn{4}{c}{\textbf{Radiology Metrics}}  \\
    \cline{ 2-9 }
                & BLEU-1 $\uparrow$ &BLEU-4 $\uparrow$ & ROUGE $\uparrow$ & BERTScore $\uparrow$ &RadCliQ $\downarrow$ & RadGraphF1 $\uparrow$ & CheXbertF1 $\uparrow$ &  GREEN $\uparrow$  \\
    \hline
    CheXbert  & \textbf{0.487} & \textbf{0.199} & \textbf{0.352} & \textbf{0.898} & \textbf{2.39} & \textbf{0.304} & \textbf{0.608} & \textbf{0.394}              \\
    \hline
    Gemini-2.0-flash  & 0.471 & 0.180 & 0.342 & 0.865 & 2.57 & 0.272 & 0.592& 0.374\\
    \hline
    GPT-5	  & 0.485 & 0.193 & 0.350 & 0.886 & 2.40 & 0.296 & 0.605 & 0.391                   \\
    \hline
  \end{tabular}}
  \caption{\label{tab7}
            Influence of different disease-status extractor on the MIMIC-CXR test set. The best results are highlighted in bold.
  }
\end{table*}
\begin{table*}
  \centering
  \scalebox{0.75}{
  \begin{tabular}{l|cccc|cccc}
    \hline
    \multirow{2}{*}{\textbf{Model}} &\multicolumn{4}{c|}{\textbf{Lexical Metrics}}  & \multicolumn{4}{c}{\textbf{Radiology Metrics}}  \\
    \cline{ 2-9 }
                & BLEU-1 $\uparrow$ &BLEU-4 $\uparrow$ & ROUGE $\uparrow$ & BERTScore $\uparrow$ &RadCliQ $\downarrow$ & RadGraphF1 $\uparrow$ & CheXbertF1 $\uparrow$ &  GREEN $\uparrow$  \\
    \hline
    w LLM-refinement & \textbf{0.487} & \textbf{0.199} & \textbf{0.352} & \textbf{0.898} & \textbf{2.39} & \textbf{0.304} & \textbf{0.608} & \textbf{0.394}              \\
    \hline
     wo LLM-refinement & 0.484 & 0.196 & 0.349 & 0.895 & 2.41 & 0.301 & 0.608& 0.393 \\
    \hline
  \end{tabular}}
  \caption{\label{tab8}
            Influence of LLM refinement on the MIMIC-CXR test set. The best results are highlighted in bold.
            }
\end{table*}

\subsection{Evaluation of Different Disease-status Extractor}
CheXbert is a widely used disease-status extractor for CE-metric evaluation, and we adopt it for disease extraction and TP/FN/FP partitioning. To quantify potential error propagation, we evaluated 1,000 samples from the MIMIC-CXR test set by comparing CheXbert-extracted disease statuses against the ground-truth labels, and observed 98.9\% extraction accuracy, suggesting that extractor noise is limited. Additionally, we also measured the extraction accuracy of Gemini-2.0-flash and GPT-5, which achieve 89.3\% and 92.1\%, respectively. As a sensitivity check, we replaced CheXbert with Gemini-2.0-flash or GPT-5 for disease-status extraction and CE-metric evaluation during the overall training and inference stage. As shown in the Table \ref{tab7}, CheXbert delivers the best overall performance, while Gemini-2.0-flash or GPT-5 incur additional inference cost without improving results. This may be because CheXbert incorporates domain-specific knowledge. Overall, these findings indicate that our conclusions are robust to the choice of extractor and highlight a clear accuracy–compute trade-off.

\subsection{Influence of LLM Refinement}
To disentangle the contributions of ESC-RL versus LLM refinement, we evaluate report quality w/w LLM (referred as GPT-5) refinement. As shown in the table below, radiology and CE metrics remain largely unchanged, while lexical metrics improve noticeably, indicating that the LLM primarily enhances fluency and readability rather than core clinical diagnosis. Therefore, the gains are mainly driven by ESC-RL, while the framework can further benefit from integration with more capable LLMs.

\section{Prompt Design for Observation Re-integration}
We provide the prompt template used in the ESC-RL framework to re-integrate multiple clinical observations into a refined radiology report. The prompt guides LLMs to filter noisy or unreliable descriptions based on trustworthy disease labels, retain only evidence-consistent findings, and correct factual inconsistencies without introducing new information. By explicitly constraining the output to trusted disease evidence, the prompt facilitates self-correcting report refinement and improves clinical consistency.
\begin{figure*}
\centering
\includegraphics[width=0.95\textwidth]{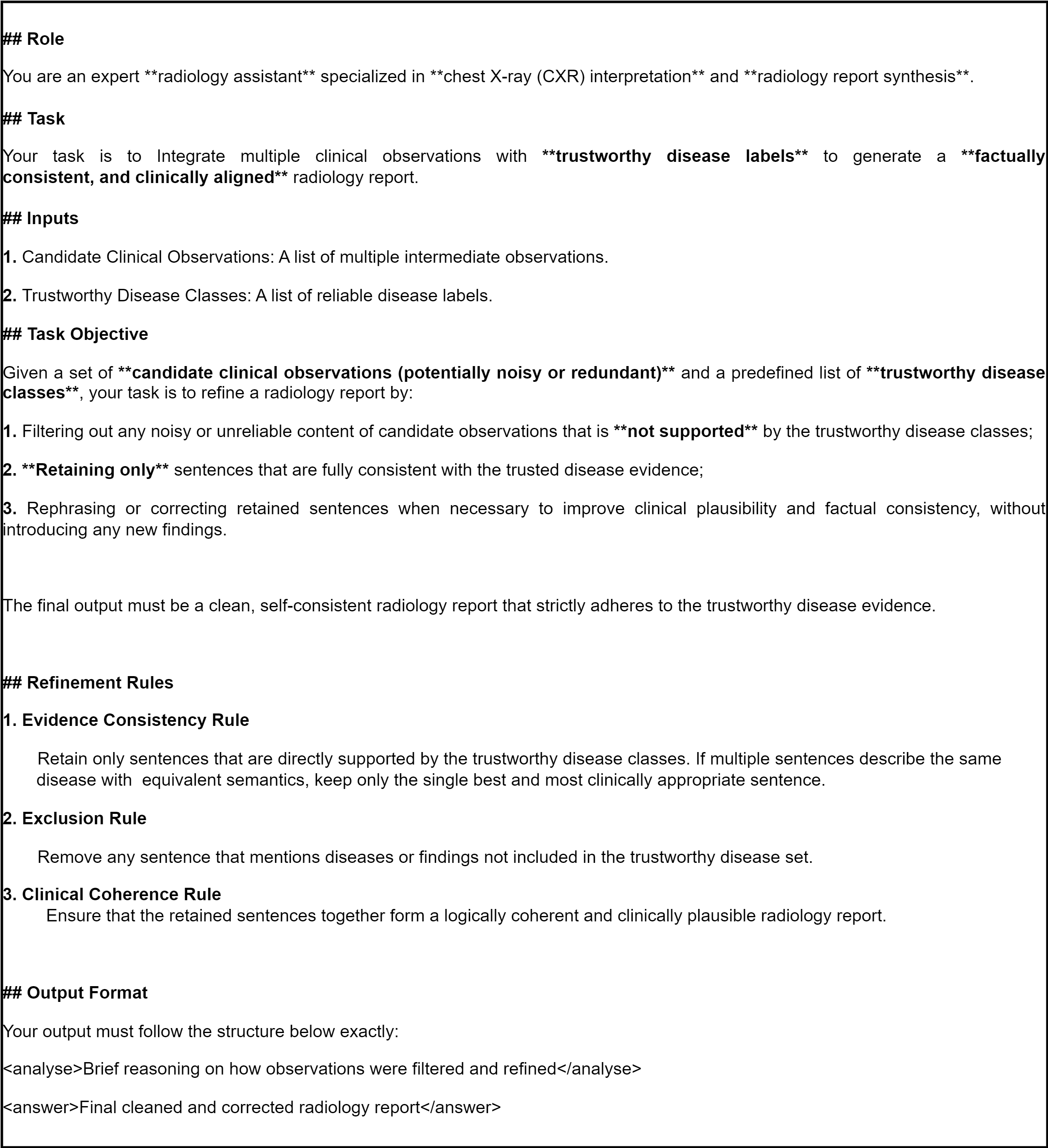}
\caption{The prompt template used for re-integrating multiple observations into a refined report within the ESC-RL framework using LLMs.}
\label{fig6:env}
\end{figure*}

\end{document}